\documentclass[11pt,letterpaper]{article}
\usepackage{ijcnlp2017}
\usepackage{times}
\usepackage{latexsym}
\usepackage[ruled,vlined,boxed]{algorithm2e}
\usepackage{graphicx}
\usepackage{url}
\usepackage{amsmath}
\usepackage{verbatim}
\usepackage{array}

\ijcnlpfinalcopy

\def\ijcnlppaperid{***}

\title{Improving Text Normalization by Optimizing Nearest Neighbor Matching}

\author{Salman Ahmad Ansari \and Usman Zafar \and Asim Karim \\
  {\tt \{15030013, 14030017, akarim\}@lums.edu.pk}}

\date{}

\begin{document}

\maketitle
\vspace{-5ex}

\begin{abstract}
\vspace{-1ex} Text normalization is an essential task in the processing and analysis of social media that is dominated with informal writing. It aims to map informal words to their intended standard forms. In this paper, we present an automatic optimization-based nearest neighbor matching approach for text normalization. This approach is motivated by the observation that text normalization is essentially a matching problem and nearest neighbor matching with an adaptive similarity function is the most direct procedure for it. Our similarity function incorporates weighted contributions of contextual, string, and phonetic similarity, and the nearest neighbor matching involves a minimum similarity threshold. These four parameters are tuned efficiently using grid search. Our approach matches each out-of-vocabulary (OOV) word to its intended in-vocabulary (IV) word rather than the usual approach of candidate-generation-and-filtering of OOV words for each IV word.  We evaluate the performance of our approach on two benchmark datasets. The results demonstrate that parameter tuning on small sized labeled datasets produce state-of-the-art text normalization performances.  
\end{abstract}
\section{Introduction}
\vspace{-1ex}Social media has proliferated the use of informal writings. In particular, such writings contain numerous spelling variations that do not appear in standard lexicons. Analysis of such informal content produces poor results if unnormalized.

Text normalization is the task of mapping an out-of-vocabulary (OOV, not in lexicon) word to an in-vocabulary (IV, in-lexicon) word that best captures its intent in the writing. For example, \emph{sum1} and \emph{r} are typical informal variants of lexicon entries `someone' and `are', respectively. Two key elements are required in text normalization: (1) assessment of the relatedness or the similarity between OOV and IV words, and (2) procedure for mapping each OOV word to its intended IV word(s). The similarity between an OOV and an IV word can be quantified by relating the contexts in which these words occur in a corpus, by considering their phonetic similarity, and by determining their string similarity. While these ideas have been utilized in previous works on text normalization, the common practice is to either adopt a single idea/cue or to combine them in equal proportions to define the similarity between words (e.g., \citet{sonmez2014graph}).
Additionally, procedures for finding OOV words similar to an IV word that follow a hierarchical candidate generation and filtering approach (e.g., \citet{han2013lexical}) involve a number of user-selectable parameters that are difficult to tune for improved performance and often result in poor recall. 

In this work, we present a nearest neighbor matching approach for text normalization in which the weighted contribution of contextual, phonetic, and string similarity to the overall similarity and the matching threshold is optimized. The $K \geq 1$ nearest IV neighbors  of an OOV word with similarity greater than or equal to the threshold define the mappings. Another contribution has been the realization that mapping of OOV words to IV words instead of the traditional method of generating and filtering OOV words from IV words results in a notable increase in precision. This simple procedure produces best matches for any $K$. The weights and the threshold are optimized on a sample labeled dataset to improve normalization accuracy. We evaluate our approach on two benchmark datasets of text normalization. The results show significant improvement in precision and F-measure even when small labeled datasets are used for optimization.

\section{Related Work}
\label{sc:rel_work}
\vspace{-1ex}Typical works on text normalization have relied primarily upon string and phonetic similarity in an hierarchical candidate generation and filtering procedure for identifying lexical variations of IV words (\citet{elmagarmid2007duplicate, contractor2010unsupervised, gouws2011unsupervised, han2013lexical, ahmed2015lexical}). For example, \cite{han2013lexical} present  a technique that generates a `confusion set' for IV words by filtering out OOV words using edit distance and phonetic measures, followed by ranking based on a tri-gram language model. 

Recently, more emphasis has been placed  on sophisticated contextual information for text normalization. \citet{levy2014neural} evaluate the performance of Word2Vec embeddings on different NLP tasks including word similarity and word analogy. \citet{sridhar2015unsupervised} propose a candidate generation and filtering approach that uses word-embeddings to create the confusion set for IV words. \citet{hassan2013social} construct a bi-partite graph with context nodes on one side and word nodes on the other. Edges between words and contexts are weighted with contextual information, and Markov random walks are used to discover OOV-IV pairs. \citet{sonmez2014graph} present a word association graph that encodes the position of each word with respect to other words. Edges indicate contextual association and their weights are assigned based on co-occurrences of the words. A number of context focused learning procedures are also presented as part of a text normalization challenge (\citet{baldwin2015shared}). Our work explores a direct matching approach with an adaptive similarity function combining different notions of similarity for improved text normalization performance. 

\section{Optimized Nearest Neighbor Matching}
\label{sc:method}
\vspace{-1ex}Our proposed optimized nearest neighbor approach for text normalization is motivated by two intuitions. First, the relative contributions of different notions of similarity towards the overall similarity between words can be different for different languages and contexts (e.g., geographical regions, topics, etc). Second, nearest neighbor matching of an OOV word to IV word(s) is a direct and optimal matching strategy with few user tunable parameters. We start the discussion of our approach by formally defining the text normalization problem.

Let $\mathcal{I}$ and $\mathcal{O}$ be the set of in-vocabulary (IV) and out-of-vocabulary (OOV) words, respectively. Then, text normalization is the task of matching  each OOV word $o_j \in \mathcal{O}$ ($j = 1, \ldots, |\mathcal{O} |$) to one or more IV words $i_k \in \mathcal{I}$ ($k = 1, \ldots, |\mathcal{I}|$) such that $o_j$ is an informally spelled variant of $i_k$ in the context of interest. Text normalization is evaluated using precision, recall, and F-measure computed over a labeled dataset (i.e., a data containing correct $o_j \leftrightarrow i_k$ mappings). 

\subsection{Weighted Similarity Function}
In order to match OOV words to their most relevant IV words, we need to define relatedness or similarity between OOV and IV words. In previous works of text normalization, similarity has been defined in terms of contextual, phonetic (sound-based), and string similarity. In this work, we propose a weighted average of all three notions of similarity. Mathematically, the (overall) similarity between $o_j$ and $i_k$ is defined as
\[
S(o_j, i_k) = \frac{\sum_{i \in \{c, p, s\}} w_i S_i(o_j, i_k) } {w_c + w_p + w_s}
\]
\noindent where $S_c(\cdot, \cdot)$, $S_p(\cdot, \cdot)$, and $S_s(\cdot, \cdot)$ respectively give the contextual, phonetic, and string similarity between words, and $w_c$, $w_p$, $w_s$ are the corresponding weights. Without loss of generality, we assume all similarities and weights lie in the interval $[0,1]$. This ensures that the overall similarity $S(\cdot, \cdot)$ also lies in the interval $[0,1]$ with higher values signifying greater similarity between the words. If a similarity is not defined (e.g., contextual similarty is not known) then the corresponding weight is set to zero in the computation. 

The contextual similarity $S_c(o_j, i_k)$ between words $o_j$ and $i_k$ quantifies the relatedness of these words based on their contextual usage. A commonly-used representation of words based on their contextual usage in a corpus is provided by Word2Vec (\citet{mikolov2013efficient}). Let $\mathbf{o}_j$ and $\mathbf{i}_k$ be the learned vector representations of words $o_j$ and $i_k$, respectively. Then, their contextual similarity is defined as
\[
S_c(o_j, i_k) = \frac{ \mathbf{o}_j^{\mathrm{T}} \mathbf{i}_k} {||\mathbf{o}_j || ||\mathbf{i}_k|| }.
\]
\noindent This represents the cosine similarity between the two vectors, and it lies in the interval $[0,1]$. 

The phonetic similarity $S_p(o_j, i_k)$ between words $o_j$ and $i_k$ measures the degree of similarity in pronunciation/sound of the words. Different sound-based encoding schemes are available for different languages. For the English language, Double Metaphone has been shown to be accurate (\citet{philips2000double}). However, we do not match the encoding of similar words directly but instead we calculate the string similarity (see next paragraph) between the encodings.  Hence, based on this encoding scheme, the phonetic similarity varies between $S_p(o_j, i_k) = 1$ when any encoding matches exactly for the two words and $S_p(o_j, i_k) = 0$ when no matches are found. 

The string similarity $S_s(o_j, i_k)$ between words $o_j$ and $i_k$ quantifies the string similarity between the words. There are a number of string similarity measures available. In this work, we adopt the normalized longest common subsequence measure defined as (\citet{yujian2007normalized}) 
\[
S_s(o_j, i_k) = \frac{\mathit{lcs}(o_j, i_k) } {\min[\mathit{len}(o_j), \mathit{len}(i_k)] + \mathit{ld}(o_j, i_k) }.
\]
\noindent Here, $\mathit{lcs}(\cdot, \cdot)$ denotes the length of the longest common subsequence of the words, $\mathit{len}(\cdot)$ is the character-length of the word, and $\mathit{ld}(\cdot, \cdot)$ gives the Levenshtein distance between the words. This similarity also lies in the interval $[0,1]$. 

\begin{figure}[tb]
\centering
\includegraphics[width=0.8\linewidth]{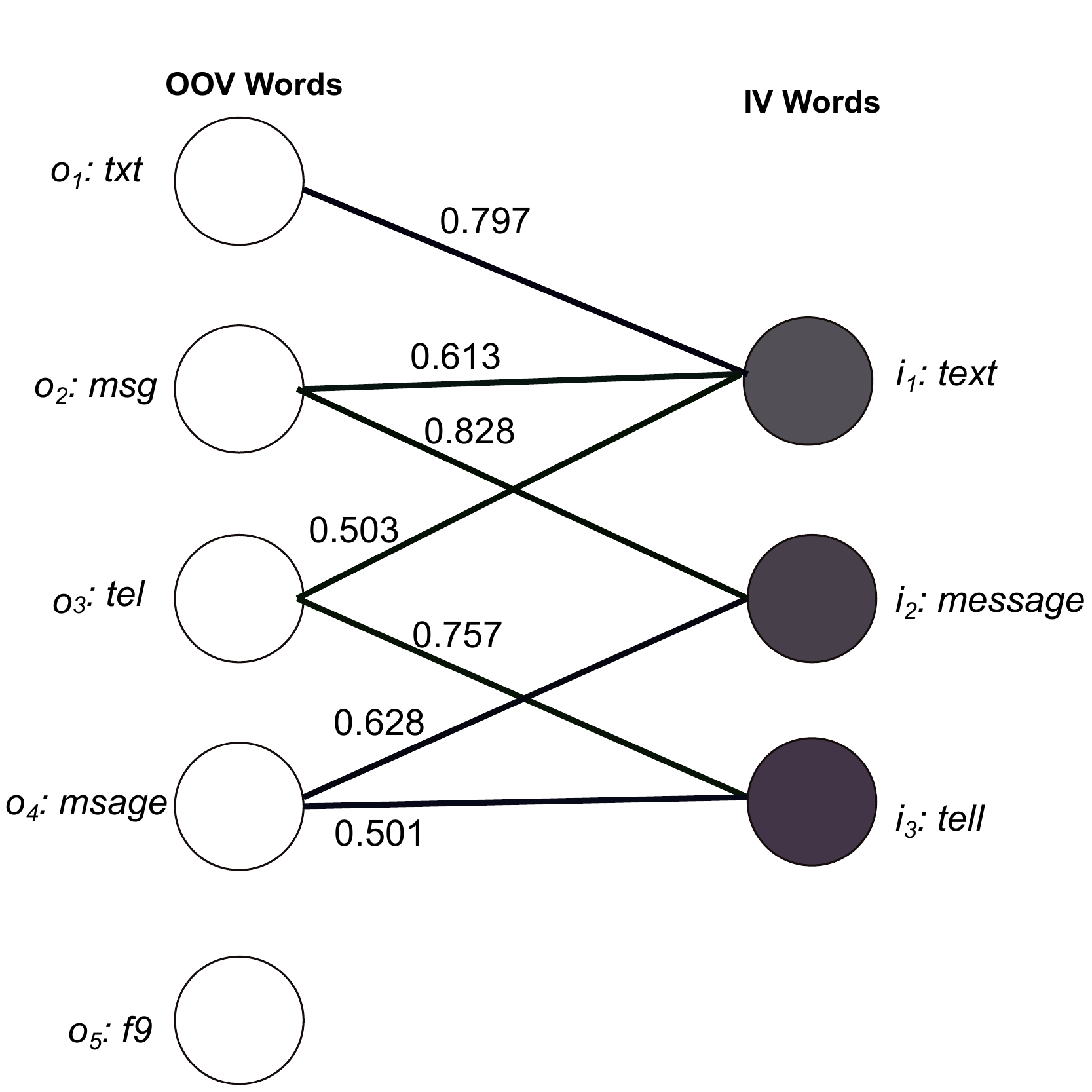}
\vspace{-2ex}
\caption{Illustration of nearest neighbor matching as bi-partite graph matching}
\label{fg:matching}
\vspace{-3ex}
\end{figure}

\subsection{Nearest Neighbor Matching}
Text normalization is essentially a matching problem. For one-to-one or one-to-many matching, the $K$ nearest neighbors approach is the most direct and appropriate. When $K=1$ each OOV word is matched with its most similar IV word, i.e., $o \leftrightarrow i_k$ when $S(o, i_k)$ is a maximum for all $k$. In addition to this standard procedure, we also introduce a minimum similarity threshold $0 < t < 1$ such that a match only occurs when the maximum similarity is greater than or equal to $t$, i.e., $o \leftrightarrow i_k$ when $S(o, i_k)$ is the highest similarity \emph{and} $S(o, i_k)\geq t$. In general, each OOV word can be matched with $K>1$ IV words that represent the top-$K$ most similar IV words of the OOV word. 

Figure \ref{fg:matching} illustrates $K$ nearest neighbor matching as a bi-partite graph matching problem. One set of nodes in the graph represents OOV words and the other set represents IV words. An edge exists between an OOV word and an IV word if their similarity is $\geq t$ and it is among the top-$K$ highest similarities of the OOV word, The figure shows edges that satisfy these conditions. It is possible that an OOV word does not have any edges, indicating that its similarity to all IV words is below the threshold $t$; such words will not be matched.

It is easy to see that given a similarity function, nearest neighbor matching (or bi-partite graph matching) will yield an optimal matching in the sense that the sum of similarities of matched OOV words is a maximum. 

\subsection{Optimizing the Parameters}
The weights in the overall similarity function control the relative contribution of each notion of similarity towards the overall similarity between words. Since contextual, phonetic, and string similarity are computed independently of the text normalization problem, it is expected that tuning these weights would improve text normalization performance. Similarly, the threshold $t$ controls the matching of OOV and IV words. 

We propose an  optimization approach for automatically tuning these parameters ($w_c$, $w_p$, $w_s$, $t$). Assuming a labeled dataset is available for training, the optimal parameter values can be found by grid search in parameter space (\citet{bergstra2012random}). This is computationally efficient since only four grid values in $[0,1] \times [0,1] \times [0,1] \times [0.1, 0.9]$ space need to be searched. For example, if jumps of $0.1$ are taken then less than 10,000 evaluation iterations are required. Note that the evaluation simply involves nearest neighbor matching and computation of text normalization performance. The grid search, which can be repeated hierarchically for refined estimates, will yield optimal parameter values ($w_c^*$, $w_p^*$, $w_s^*$, $t^*$) to be used for future text normalization. 

\section{Experimental Evaluation}
\label{sc:exp}
\vspace{-1ex}
\subsection{Datasets and Evaluation Settings}
We use a Twitter dataset with over 4.5 million tweets for learning the contextual representation of words. Based on Python's Enchant  English dictionary, this dataset has 36,071 distinct IV and 46,480 distinct OOV words. For evaluating our approach, we use two publicly available labeled datasets for text normalization: \citet{han2011lexical}\footnote{https://tq010or.github.io/research.html} (Han) and \citet{liu2011insertion} (Yang) datasets. The Han dataset contains 807 IV and 975 OOV words out of which 268 IV and 386 OOV words overlap with our Twitter dataset. The Yang dataset contains 1,975 IV and 3,937 OOV words out of which 1,030 IV and 2,141 OOV words overlap with our Twitter dataset. Our evaluation, however, is done over all the words in the datasets since we ignore contextual similarity (i.e., $w_c^* = 0$) in the overall similarity of word pairs whose contextual similarity is not defined. These datasets provide only one mapping per OOV word; hence, we perform $K=1$ nearest neighbor matching. 

We adopt standard Precision (Pre), Recall (Rec), and F-measure (Fme) to evaluate the performance of text normalization (\citet{sonmez2014graph}). 

\begin{figure}[t!]
\centering
\includegraphics[width=0.8\linewidth]{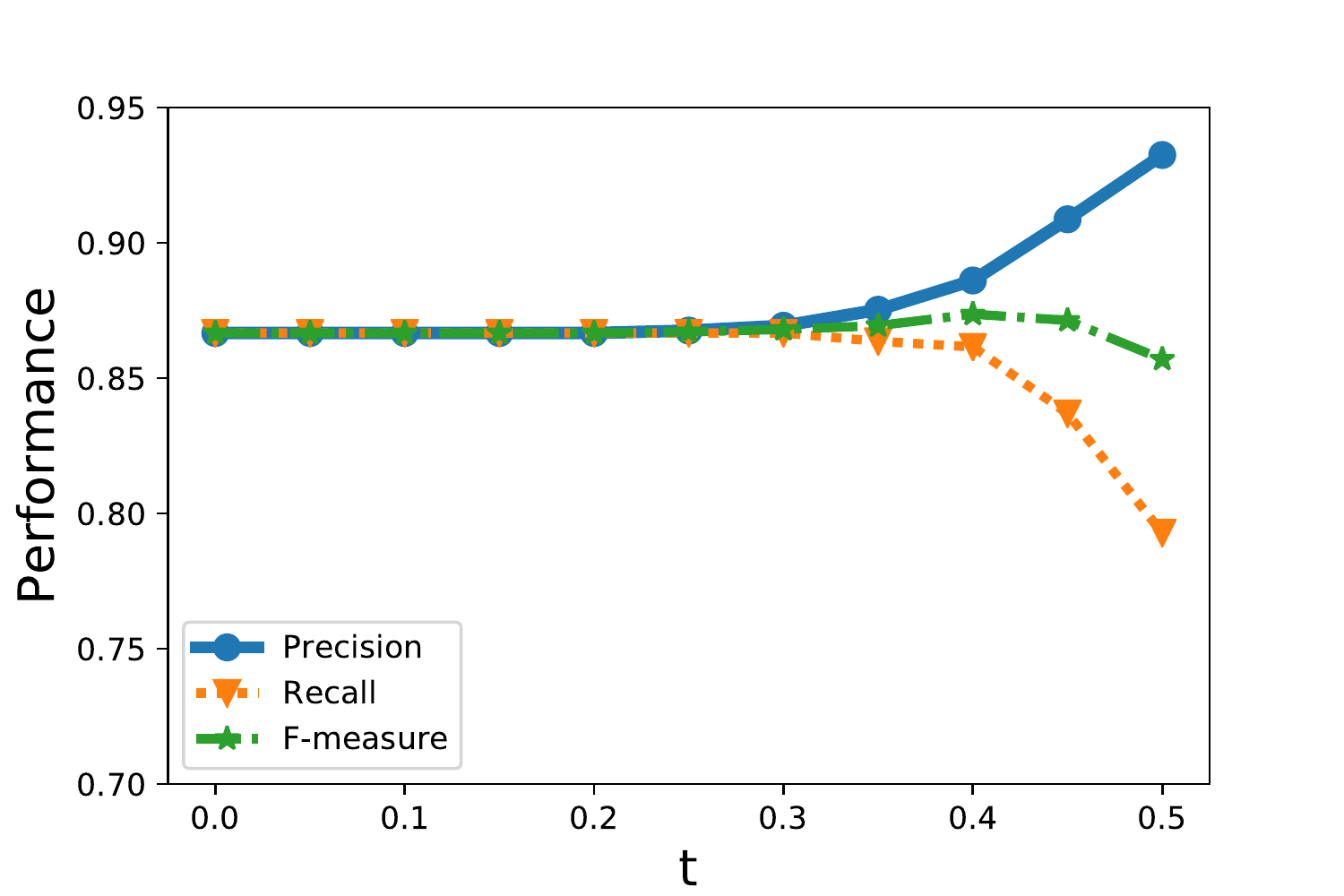}
\vspace{-2ex}
\caption{Performance with varying threshold $t$ when $w_c=1$,$w_p=0$, and $w_s=1$ on Han dataset}
\label{fg:han_trend}
\vspace{-3ex}
\end{figure}

\begin{table}[!tbh]
\small
\centering
\begin{tabular}{ |c|c|c|c|c| }
\hline
\textbf{Exp} & \textbf{$w_c^*$, $w_p^*$, $w_s^*$, $t^*$} & \textbf{Pre}&\textbf{Rec}&\textbf{Fme}\\
\hline
\multicolumn{5}{|c|}{\textbf{ Han Dataset}}\\
\hline
1& $0.9, 0.4, 0.7, 0.4$ &90.2&89.2&89.7\\
\hline
2&$0.7, 0.3, 0.5, 0.2$ &87.0&87.0&87.0\\
\hline
3&$0.7, 0.3, 0.7, 0.3$ &89.3&89.0&89.2\\
\hline
\multicolumn{5}{|c|}{\textbf {Yang's Dataset}}\\
\hline
1& $0.7, 0.3, 0.7, 0.3$ &76.1&76.0&76.0\\
\hline
2& $0.8,0.4,0.7,0.2$ &74.2&76.2&75.2\\
\hline
3&$0.9,0.4,0.7,0.4$&76.6&74.6&75.6\\
\hline
\end{tabular}
\vspace{-1ex}
\caption{Performance of our approach on Han and Yang datasets (see text for description of exp)}
\label{tb:exp}
\vspace{-3ex}
\end{table}

\subsection{Results}
We start by presenting the performance trend of our approach with varying values of threshold $t$ while weights are not optimized.  Figure \ref{fg:han_trend} shows this trend for Han dataset when  $w_c = 1$, $w_p = 0$, and $w_s = 1$ (this is the best un-optimized combination). It is seen that precision and recall remain high at low threshold values, but recall starts decreasing rapidly at higher threshold values. The highest F-measure of 87.4\% is obtained at $t = 0.4$.  

Subsequently, we conduct three different experiments to evaluate our parameter-optimized approach. Experiment 1 reports average parameter values and performance over entire dataset via 2-fold cross-validation. Experiment 2 reports average parameter values and average performance for 5 randomized runs testing over 80\% of the dataset after optimizing over the respective remaining 20\% of the dataset. Experiment 3  reports parameter values and performance over the dataset when parameters are optimized over the other dataset (e.g., performance over Han dataset after learning over Yang dataset). In Experiments 1 and 2 there is no overlap between words in the training and test samples. 

Table \ref{tb:exp} gives the results of these experiments. The highest F-measure of 89.7\% and 76\% on Han and Yang datasets, respectively, is obtained in Experiment 1. Even when a small sample size of 20\% is used for tuning the parameters (Exp 2), our approach produces F-measures of 87.0\% and 75.2\% on Han and Yang datasets, respectively. These results compare with the previous best F-measures of 82.3\% on Han dataset (\citet{sonmez2014graph}) and 73\% on Yang dataset (\citet{yang2013log}). A review of the optimal parameters reveal that contextual and string similarity play dominant roles in text normalization while phonetic similarity is less significant. These results confirm that our approach is not only simpler but also more  accurate when contextual information of words and small sized labeled dataset is available. 
\vspace{-1ex}
\section{Concluding Remarks}
\label{sc:con}
\vspace{-1ex}
We have presented and evaluated an efficient optimized nearest neighbor (NN) matching approach for improved text normalization. Results on two benchmark datasets have demonstrated that weights in similarity function and threshold in NN matching can be tuned  over small labeled samples to yield state-of-the-art performance.


\bibliography{ijcnlp2017}
\bibliographystyle{ijcnlp2017}

\end{document}